%% file: main.tex
\documentclass[10pt,twocolumn,letterpaper]{article}
\usepackage[pagenumbers]{cvpr} %
\input{preamble}
\definecolor{cvprblue}{rgb}{0.21,0.49,0.74}
\usepackage[pagebackref,breaklinks,colorlinks,citecolor=cvprblue]{hyperref}

\newcommand{\SMa}{Appendix~\ref{app:implementation_details}}
\newcommand{\SMb}{Appendix~\ref{app:benchmark_details}}
\newcommand{\SMc}{Appendix~\ref{app:negative}}
\newcommand{\SMd}{\textit{We have additional analysis in our Appendix, including a detailed analysis on weight interference (Appendix~\ref{app:interference_analysis}), plasticity and KID~\cite{binkowski2018demystifying} score metrics (Appendix~\ref{app:metrics}), a plot of plasticity vs number of trained tasks (\ref{app:plastic_analysis}), and variance across runs (Appendix~\ref{app:variance}).}}
\newcommand{\SMe}{Appendix~\ref{app:details_classification}}
\newcommand{\SMf}{See \ref{appendix:source} for source of target images.}

\begin{document}

\title{Continual Diffusion with STAMINA:\\STack-And-Mask INcremental Adapters}

\author{\textbf{James Seale Smith\textsuperscript{1,2} \quad Yen-Chang Hsu\textsuperscript{1} \quad Zsolt Kira\textsuperscript{2} \quad Yilin Shen\textsuperscript{1} \quad Hongxia Jin\textsuperscript{1}} 
\\
\normalsize
\textsuperscript{1}Samsung Research America,
\textsuperscript{2}Georgia Institute of Technology
}

\input{figures/teaser}

\thispagestyle{empty}

\input{sections/0_abstract}

\input{sections/1_intro.tex}
\input{sections/2_related.tex}
\input{sections/3_prelim}

\input{sections/4_method}
\input{sections/5_experiments}
\input{sections/6_conclusions}

\bibliographystyle{unsrt}
\bibliography{references}

\clearpage
\section*{Appendix}
\setcounter{figure}{0}
\setcounter{table}{0}
\renewcommand{\thetable}{\Alph{table}}
\renewcommand{\thefigure}{\Alph{figure}}
\renewcommand\thesection{\Alph{section}}
\appendix
\input{sections/7_appendix_arxiv}

\end{document}

%% file: preamble.tex
\usepackage{graphicx}
\usepackage{amsmath}
\usepackage{amssymb}
\usepackage{booktabs}
\usepackage{xcolor}
\usepackage{array,makecell}
\usepackage{caption}
\usepackage{enumitem}
\usepackage{amsmath}
\usepackage{bm}
\usepackage{arydshln}
\usepackage[accsupp]{axessibility}
\usepackage{multirow}

\usepackage{subcaption}
\usepackage{wrapfig,lipsum}
\newcommand {\red}[1]{{\color{purple}\textbf{#1}}\normalfont}
\newcommand {\blue}[1]{{\color{blue}\textbf{#1}}\normalfont}
\newcommand{\vl}{{VL}}
\def\setting{{Continual Diffusion}}
\def\method{{STAMINA}}

%% file: figures/teaser.tex
\twocolumn[{%
\renewcommand\twocolumn[1][]{#1}%
\maketitle

\begin{center}
    \centering
    \vspace{-2mm}
    \includegraphics[width=\textwidth]{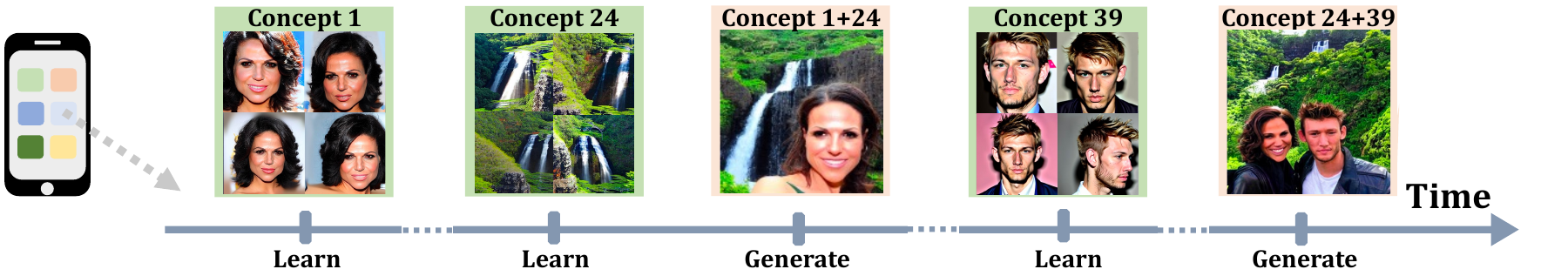}
    \captionof{figure}{
    Our work demonstrates sequentially learning long sequences of concepts. At any time, we can generate photos of any prior learned concepts, including multiple concepts together. \emph{Images denoted as ``generate'' in this figure are real results from our method.}
    }
    \vspace{3mm}
    \label{fig:use-case}
\end{center}
}]

%% file: sections/0_abstract.tex
\begin{abstract}
    \vspace{-5mm}
    Recent work has demonstrated a remarkable ability to customize text-to-image diffusion models to multiple, fine-grained concepts in a sequential (i.e., continual) manner while only providing a few example images for each concept. This setting is known as \emph{continual diffusion}. Here, we ask the question: \emph{Can we scale these methods to longer concept sequences without forgetting?} Although prior work mitigates the forgetting of previously learned concepts, we show that its capacity to learn new tasks reaches saturation over longer sequences. We address this challenge by introducing a novel method, STack-And-Mask INcremental Adapters (STAMINA), which is composed of low-ranked attention-masked adapters and customized MLP tokens. STAMINA is designed to enhance the robust fine-tuning properties of LoRA for sequential concept learning via learnable hard-attention masks parameterized with low rank MLPs, enabling precise, scalable learning via sparse adaptation. Notably, all introduced trainable parameters can be folded back into the model after training, inducing no additional inference parameter costs. We show that STAMINA outperforms the prior SOTA for the setting of text-to-image continual customization on a 50-concept benchmark composed of landmarks and human faces, with no stored replay data. Additionally, we extended our method to the setting of continual learning for image classification, demonstrating that our gains also translate to state-of-the-art performance in this standard benchmark.
\end{abstract}
\vspace{-1mm}

%% file: sections/1_intro.tex
\vspace{-.55cm}
\section{Introduction}
\label{sec:intro}

\looseness=-1
Remarkable progress in text-to-image diffusion models has ushered in an era of practical applications ranging from generating detailed product images for e-commerce and advertising platforms to facilitating creative endeavors in content generation and story-telling. A particularly compelling direction in this field is the task of customizing models to multiple, fine-grained concepts in a \emph{continual} manner, a recently introduced paradigm known as \setting~\cite{smith2023continual}. In this setting, models are sequentially adapted to new concepts using a few example images without forgetting prior learned concepts. This is an import problem for practical use cases such as customizing a generative model for a user over time without requiring permissions to store their private and personal user data (visualized in Figure~\ref{fig:use-case}). While prior work has shown significant strides in mitigating catastrophic forgetting for this setting, we ask: \emph{How does the model's performance evolve when we scale to longer concept sequences?}

In this paper, we start by addressing this question and demonstrate with quantitative analysis that, while the state-of-the-art (SOTA) method for this task (C-LoRA~\cite{smith2023continual}) can significantly alleviate catastrophic forgetting, the capacity to learn new tasks rapidly saturates after a certain number of tasks. In response, we propose an innovative approach, \textbf{STack-And-Mask INcremental Adapters (STAMINA)}, which combines the robustness of low-rank adaptations with the precision of attention masking, and enhances them with learnable MLP tokens. STAMINA comprises two key elements: 1) Low-Rank Adapters (LoRA) with hard-attention masks parameterized by low-rank MLP modules and Gumbel softmax, and 2) learnable MLPs that replace the custom token feature embeddings used in prior works~\cite{gal2022image,kumari2022multi,smith2023continual}. Importantly, all trainable parameters introduced can be seamlessly folded back into the model post-training, thereby not inducing any additional parameter costs during inference. Our method exceeds the prowess of prior C-LoRA for \setting~by strongly surpassing its plasticity to learn longer concept sequences without sacrificing its resilience to catastrophic forgetting.

We demonstrate that STAMINA significantly outperforms the prior SOTA in text-to-image continual customization without any stored replay data on 3 benchmarks, including a comprehensive 50-concept benchmark composed of human faces and landmarks. Moreover, we show that STAMINA achieves this performance while requiring significantly fewer training steps compared to the previous SOTA. \textbf{In summary, our contributions in this paper are as follows:}
\begin{enumerate}
[topsep=0pt,itemsep=-1ex,partopsep=1ex,parsep=1ex,leftmargin=*,labelindent=0pt]
\item We empirically demonstrate the limitations of existing approaches in scaling to longer task sequences in the \setting~setting.
\item We propose the STAMINA method, a novel combination of LoRA with attention masking and MLPs, which boosts the model's ability to learn and remember longer sequences of tasks. Importantly, all of the trainable parameters introduced by STAMINA can be integrated back into the model post-training, ensuring no additional inference parameter costs.
\item We present extensive experiments showing that STAMINA significantly outperforms the current state-of-the-art in the \setting~setting on a 50-concept benchmark while requiring significantly fewer training steps.
\item We demonstrate the versatility of \method~ and extend it into the setting of continual learning for image classification, and show that our gains translate to state-of-the-art performance for a widely used 20-task benchmark.
\end{enumerate}

%% file: sections/2_related.tex
\section{Background and Related Work}
\label{sec:rl}

\looseness=-1
\textbf{Conditional Image Generation}: Conditional image generation is a well-studied topic with several approaches, including Generative Adversarial Networks (GANs)\cite{goodfellow2014generative,Karras_2020_CVPR}, Variational Autoencoder (VAE)\cite{kingma2013auto}, and diffusion models~\cite{dhariwal2021diffusion,ho2020denoising,sohl2015deep}. We focus on the popular diffusion-based models that use free-form text prompts as conditions \cite{ramesh2022hierarchical, liu2022compositional}. These models operate by iteratively adding noise to the original image and then removing the noise through a backward pass to produce a final generated image. To inject text conditions, a cross-attention mechanism is introduced in the transformer-based U-Net~\cite{ronneberger2015u}.

Recent works have explored the generation of \emph{custom} concepts, such as a unique stuffed toy or a pet. Dreambooth~\cite{ruiz2022dreambooth} fine-tunes the whole parameter set in a diffusion model using given images of the new concept, while Textual Inversion~\cite{gal2022image} learns custom feature embedding ``words''.  Methods are not limited to conditioning on only text - recent work has shown that these models can be customized to effectively add conditioning with new modalities (such as semantic segmentation maps and sketches)~\cite{ham2023modulating}. In contrast, Custom Diffusion~\cite{kumari2022multi} learns multiple concepts using a combination of cross-attention fine-tuning, regularization, and closed-form weight merging. However, Custom Diffusion struggles to learn similar, fine-grained concepts in a sequential manner (i.e., \setting), motivating the recent C-LoRA~\cite{smith2023continual} work which is the first to sequentially customize Stable Diffusion in a ``continual learning'' manner. We mention that several recent pre-print works~\cite{gal2023designing,han2023svdiff,liu2023cones,shi2023instantbooth,wei2023elite} have emerged that build or extend upon Dreambooth and Custom Diffusion, but none of these methods are designed for the sequential setting, and the contributions can be considered as orthogonal to our paper.

\textbf{Continual Learning}: 
Continual learning involves training a model on a sequence of tasks, each with a different data distribution, while preserving the knowledge learned from previous tasks. Existing methods to address the issue of catastrophic forgetting can be categorized into three groups~\cite{mccloskey1989catastrophic}. Regularization-based methods~\cite{douillard2020podnet, kirkpatrick2017overcoming, li2016learning, zenke2017continual} add regularization terms to the objective function during training a new task. For instance, EWC~\cite{kirkpatrick2017overcoming} estimates the significance of model parameters and applies per-parameter weight decay. Rehearsal-based methods~\cite{chaudhry2018efficient,chaudhry2019episodic,hou2019learning,kamra2017deep,pham2021dualnet, Rebuffi:2016,rolnick2019experience,smith2021abd,van2020brain} save or synthesize samples from past tasks in a data buffer and replay them alongside the new task's data. Nonetheless, privacy or copyright issues may prevent using this method. Architecture-based methods~\cite{aljundi2017expert,li2019learn,Rusu:2016,yoon2017lifelong} separate model parameters for each task. Recent prompt-based continual learning methods for Vision Transformers such as L2P~\cite{wang2022learning}, DualPrompt~\cite{wang2022dualprompt}, and CODA-Prompt~\cite{smith2022coda} have surpassed rehearsal-based methods in classification problems without needing a replay buffer. Although they have shown success in classification problems, their utility in text-to-image generation is unclear since they infer discriminative features of data to create prompts for classification. However, we do compare to them in their original problem setting.

While most research on continual learning focuses on uni-modal problems, a few approaches have been suggested for the multimodal setting. REMIND~\cite{hayes2020remind} proposed continual VQA tasks with latent replay but require storing compressed training data. CLiMB~\cite{climb} adapted CL to coarsely different \vl{} tasks, including VQA, NLVR, VE, and VCR, assuming knowledge of the evaluated task-id at inference time. Construct-VL~\cite{smith2022construct} concentrates on natural language visual reasoning. We formulate our problem as the continual adaptation of Stable Diffusion to multiple, fine-grained concepts, and most of the methods reviewed in this section do not directly apply to our problem. The only relevant work, C-LoRA~\cite{smith2023continual}, is discussed in the next section.

\textbf{Sparsity for Continual Learning}: A key component of our approach is \emph{sparse model adaptations} via attention masking to mitigate forgetting and interference in our model. Prior works have also leveraged sparsity for continual learning in other problem settings. HAT~\cite{serra2018overcoming} also applies hard attention masks, but these are task-conditioned masks on network paths within the model. This approach require task id during inference and thus cannot be applied to multi-concept generation, excluding them from our setting. DGM~\cite{ostapenko2019learning} applies sigmoid attention masks to increase sparsity and scalability for image classification, but this approach requires dynamic parameter expansion, which would not be practical in the setting of continually adapting a \emph{large-scale pre-trained} model. The work of Schwarz \emph{et al.}~\cite{schwarz2021powerpropagation} also demonstrates that sparsity can reduce catastrophic forgetting via a sparsity-inducing weight re-parameterization. It may be possible for this method to be implemented to work for adapting an existing pre-trained model, but that is out of the scope for our paper. GPM~\cite{abbasi2022sparsity} demonstrates positive effects for sparsity in continual learning with k-winner activation MLPs, but it is also not clear how this could be implemented in the context of adapting our pre-trained text-to-image diffusion model. Finally, sparsity has also been applied to meta-continual learning~\cite{von2021learning}, which requires additional training on a meta-dataset. While these works demonstrate the fundamental advantages for sparsity in continual image classification, it is not clear how they can be applied to our specific setting of text-to-image customization, motivating our approach.

%% file: sections/3_prelim.tex
\section{A Closer Look at \setting}
\label{sec:prelim}

In the \setting~setting, we learn $N$ customization ``tasks'' $t \in \{ 1,2,\dots,N-1,N\}$, where $N$ is the total number of concepts that will be shown to our model. The recent C-LoRA~\cite{smith2023continual} was the first to propose a method for \setting. In this section, we will first define some preliminaries and then review the C-LoRA method and discuss its shortcomings.

C-LoRA utilizes parameter-efficient fine-tuning of a transformer. Consider the context of the single-head cross-attention operation~\cite{Vaswani2017attention}, given as $\mathcal{F}_{attn}(Q,K,V) = \sigma \left(\frac{QK^{\top}}{\sqrt{d'}}\right)V$, where $\sigma$ stands for the softmax operator, $Q=\bm{W}^{Q}\bm{f}$ represents query features, $K=\bm{W}^{K}\bm{c}$ serves as key features, and $V=\bm{W}^{V}\bm{c}$ functions as value features. Additionally, $\bm{f}$ indicates latent image features, $\bm{c}$ denotes text features, and $d'$ is the output dimensionality. In this equation, the matrices $\bm{W}^{Q},\bm{W}^{K},\bm{W}^{V}$ map inputs $\bm{f}$ and $\bm{c}$ to the query, key, and value features, respectively.

Prior works~\cite{smith2023continual,kumari2022multi} only modify $\bm{W}^{K},\bm{W}^{V}$ (referred to as $\bm{W}^{K,V}$) which project the text features. In learning a customization ``task'' $t$, the following loss is minimized in the prior C-LoRA~\cite{smith2023continual}:
\begin{equation}
    \underset{\bm{W}^{K,V}_{t} \in \theta}{\operatorname{min}}\mathcal \quad {L}_{SD}(x,\bm{\theta})+ \lambda_{f} \mathcal{L}_{forget}(\bm{W}^{K,V}_{t-1},\bm{W}^{K,V}_{t}) 
\end{equation}
Here, $x$ stands for the input data of the new concept, $\mathcal{L}_{SD}$ denotes the loss function for Stable Diffusion with respect to model $\theta$, $\mathcal{L}_{forget}$ minimizes forgetting between old task $\bm{W}^{K,V}_{t-1}$ and new task $\bm{W}^{K,V}_t$, and $\lambda_{f}$ is a hyperparameter selected through a straightforward exponential sweep.

Smith \emph{et al.}~\cite{smith2023continual} propose to parameterize the weight change between old task $\bm{W}^{K,V}_{t-1}$ and new task $\bm{W}^{K,V}_{t}$ using LoRA~\cite{lora}\footnote{This was also proposed for NLVR~\cite{smith2022construct} and offline customization~\cite{lora_stable}.}, which decompose the weight matrices into low-rank residuals, as expressed by:
\begin{equation}
\begin{aligned}
    \bm{W}^{K,V}_t &= \bm{W}^{K,V}_{t-1} + \bm{A}^{K,V}_t  \bm{B}^{K,V}_t \\
                   &= \bm{W}^{K,V}_{init} + \sum_{t'=1}^{t-1}  \bm{A}^{K,V}_{t'}  \bm{B}^{K,V}_{t'}  + \bm{A}^{K,V}_t  \bm{B}^{K,V}_t
\end{aligned}\label{eq:clora}
\end{equation}

Here, $ \bm{A}^{K,V}_t \in \mathbb{R}^{D_1 \times r}$,  $\bm{B}^{K,V}_{t} \in \mathbb{R}^{r \times D_2}$, with $\bm{W}^{K,V} \in \mathbb{R}^{D_1 \times D_2}$, and $r$ being a hyper-parameter controlling the rank of the weight matrix update, chosen using a simple grid search. The initial values from the pre-trained model are represented as $\bm{W}^{K,V}_{init}$. 

C-LoRA additionally proposed a novel regularization method which involves penalizing the LoRA parameters $\bm{A}^{K,V}_{t}$ and $\bm{B}^{K,V}_{t}$ for altering locations that have been previously modified by earlier concepts in $\bm{W}^{K,V}_{t}$. Specifically, C-LoRA contains the forgetting loss $\mathcal{L}_{forget}$, given as:
\begin{equation}
\mathcal{L}_{forget} = ||\left|\sum_{t'=1}^{t-1}  \bm{A}^{K,V}_{t'}  \bm{B}^{K,V}_{t'} \right| \odot \bm{A}^{K,V}_t  \bm{B}^{K,V}_t||^2_2
\label{eq:lora-reg}
\end{equation}
where $\odot$ represents the element-wise product, or the Hadamard product.

\subsection{The Plasticity Problem}
\label{sec:plastic-problem}

\input{figures/plastic_problem}
In Section~\ref{sec:exp_vl}, we find that C-LoRA suffers in our benchmarks which involve longer task sequences. \emph{Why could that be?} Our intuition is that, as the weights diverge further from the pre-trained backbone, $\mathcal{L}_{forget}$ becomes more and more restrictive, thus limiting the ability to learn new tasks (plasticity). We show this in Figure~\ref{fig:plastic-prob} by plotting the average distance from the pre-trained weights, given as $||\bm{W}^{K,V}_{t}-\bm{W}^{K,V}_{init}||_2$, versus tasks. We see that, while the model tends to see high changes in the early tasks, this rapidly saturates, suggesting that low plasticity may be a contributor for the diminishing performance in C-LoRA on longer task sequences. These results motivate us to find a better approach for \emph{scalable} \setting.

%% file: figures/plastic_problem.tex
\begin{figure}
    \centering
    \includegraphics[width=.45\textwidth]{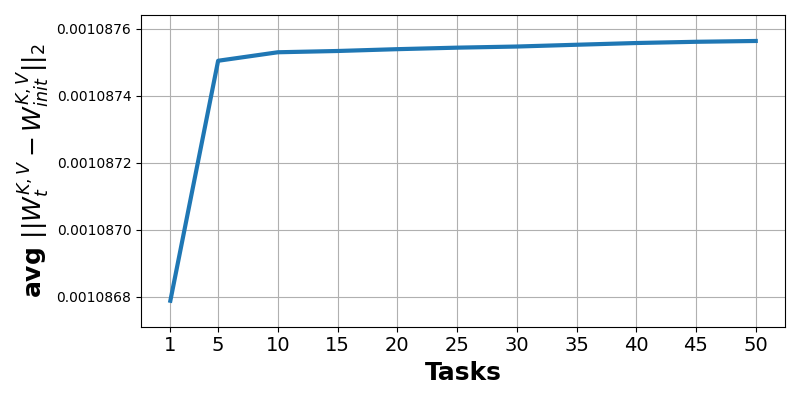}
    \caption{
    Average distance from pre-trained weights, given as $||\bm{W}^{K,V}_{t}-\bm{W}^{K,V}_{init}||_2$, vs task for C-LoRA~\cite{smith2023continual}.
    }
    \vspace{-0.2cm}
    \label{fig:plastic-prob}
\end{figure}

%% file: sections/4_method.tex
\section{The STAMINA Approach}
\label{sec:method}

Inspired by the analysis of the previous section, we propose an innovative approach to~\setting: \textbf{Stack-And-Mask INcremental Adapters (\method)}, composed of stack-able low-rank adapter with learnable hard-attention masks to encourage precise and sparse weight residuals on the pretrained diffusion model. The intuition behind our approach is that \emph{sparse} adaptations are less likely to interfere and, therefore, will avoid saturating plasticity as the number of tasks grow. The overview of our proposed method is illustrated in Figure~\ref{fig:method}, and the rest of this section motivates and describes each component of STAMINA.

\looseness=-1
\textbf{Hard Attention Masks with Gumbel-Softmax}: 
Consider the low rank weight decomposition $\bm{W}^{K,V}_t = \bm{W}^{K,V}_{t-1} + \bm{A}^{K,V}_t  \bm{B}^{K,V}_t$. The intuition is that low-rank adaptation is robust against over-fitting, making it a strong continual learner~\cite{smith2022construct,smith2023continual}. However, there is one glaring weakness with this approach: \textbf{precision}. Specifically, the low-rank properties of the product $\bm{A}^{K,V}_t  \bm{B}^{K,V}_t$ lack the ability to target important spots in the weight matrix without incurring many unnecessary adaptations to other spots in the matrix.

To reduce interference between learned concepts, we propose to apply hard-attention masks on the $\bm{A}^{K,V}_t  \bm{B}^{K,V}_t$ product. Rather than learn masks with continuous outputs on the range $ \left[0,1\right] $, as many sparse continual learning works discussed in Section~\ref{sec:rl} do, we desire a true discrete binary mask to retain desirable robust properties of the low-rank weight residuals (a decision validated by our ablations). Thus, we leverage\footnote{We note that Shen \emph{et al.}~\cite{shen2018sharp} use a similar approach in another setting.} the Gumbel-Softmax~\cite{jang2016categorical} operation, which provides a differentiable approximation to the discrete argmax operation, enabling the learning of true \emph{hard attention} masks during training.

We consider our mask as a \emph{binary categorical distribution} and expand in an additional dimension of size 2, and take the operation over the expanded dimension. With this parameterization, values of $1$ in the first index of the expanded dimension equates to ``pass through the mask'' after the Gumbel-Softmax, and the other dimension can be discarded. Specifically, we learn mask $\bm{\mathcal{M}}^{K,V}_{t}  \in \mathbb{R}^{D_1 \times D_2}$ as:
\begin{equation}
\bm{\mathcal{M}}^{K,V}_{t,i,j} = \frac{\exp\left(\frac{\log(\bm{\hat{m}}^{K,V}_{t,i,j,1})+g_{i,j,1}}{\tau}\right)}{\sum_{z=0}^{1} \exp\left(\frac{\log(\bm{\hat{m}}^{K,V}_{t,i,j,z})+g_{i,j,z}}{\tau}\right)}
\label{eq:gs_tensor}
\end{equation}
where  $\bm{\hat{m}}^{K,V}_{t}  \in \mathbb{R}^{D_1 \times D_2 \times 2} $ represents a learnable mask tensor \emph{before} Gumbel-Softmax is taken over the expanded third dimension; $i,j$ represent taking this operation over $i =1,\dots,D_1$ and $j=1,\dots,D_2$; $\tau$ is a temperature hyperparameter controlling smoothness\footnote{We use a default value of $\tau=0.5$}; and $g$ are i.i.d samples drawn from $\text{Gumbel}(0,1)$~\cite{jang2016categorical}. Notice that $\bm{\mathcal{M}}^{K,V}_{t}$ denotes our final learned mask which is applied to the $\bm{A}^{K,V}_t  \bm{B}^{K,V}_t$ product, whereas $\bm{\hat{m}}^{K,V}_{t}$ denotes the learnable parameters before Gumbel-Softmax.

Using this masking approach, our weight residuals are now given as:
\begin{equation}
\begin{aligned}
    \bm{W}^{K,V}_t &= \bm{W}^{K,V}_{t-1} + \bm{A}^{K,V}_t  \bm{B}^{K,V}_t \odot \bm{\mathcal{M}}^{K,V}_{t} \\
                   &= \bm{W}^{K,V}_{init} + \left[\sum_{t'=1}^{t-1}  \bm{A}^{K,V}_{t'}  \bm{B}^{K,V}_{t'} \odot \bm{\mathcal{M}}^{K,V}_{t'} \right] \\
                   &+ \bm{A}^{K,V}_t  \bm{B}^{K,V}_t \odot \bm{\mathcal{M}}^{K,V}_{t}
\end{aligned}\label{eq:stamina}
\end{equation}

\input{figures/method}

\textbf{MLP Mask Parameterization}: 
Rather than optimize a fixed tensor, we further enhance our masking capacity and flexibility with a low-rank multi-layer perception (MLP) parameterization, $\theta_{\bm{\mathcal{M}}^{K,V}_{t}}$, operating on a fixed input $\bm{1}$. The idea is to leverage the power of MLPs to learn more complex transformations between tasks, thereby mitigating the risk of catastrophic forgetting. Specifically, we propose to learn our mask as:
\begin{equation}
\theta_{\bm{\mathcal{M}}^{K,V}_{t,i,j}} = \frac{\exp\left(\frac{\log(\theta_{\bm{\hat{m}}^{K,V}_{t,i,j,1}})+g_{i,j,1}}{\tau}\right)}{\sum_{z=0}^{1} \exp\left(\frac{\log(\theta_{\bm{\hat{m}}^{K,V}_{t,i,j,z}})+g_{i,j,z}}{\tau}\right)}
\label{eq:gs_mlp}
\end{equation}
where $\theta_{\bm{\hat{m}}^{K,V}_{t}}$ is the learnable mask MLP \emph{before} Gumbel-Softmax is taken. As demonstrated later with our ablations, the MLP parameterization is fundamental to our method - we found simply optimizing a mask matrix directly to be ineffective as the mask would have little to no updates during learning.

\looseness=-1 In order to keep the number of learnable parameters low and not increase the search space over new hyperparameters, we leverage a very simple two layer MLP for $\theta_{\bm{\mathcal{M}}^{K,V}_{t}}$ which consists of two linear layers and ReLU~\cite{agarap2018deep} operating on the fixed input tensor $\bm{1}$. All dimensions before the final layer are of dimension $r$, the same low rank as $\bm{A}^{K,V}_{t}$ and $\bm{B}^{K,V}_{t}$. While Eq.~\eqref{eq:stamina} decomposes how the pre-trained weights are adapted, \textbf{we emphasize that all of our learned parameters can be directly folded back into the original pre-trained weight, incurring no additional storage or computation costs at inference.}

\textbf{Sparsity Regularization}: 
In order to achieve desirable \emph{sparsity} properties of the attention masks, we introduce a sparsity regularization on the positive outputs of $\theta_{\bm{\mathcal{M}}^{K,V}_{t}}$. The simple regularization encourages the mask to produce a $0$ for each spot in the weight matrix residual rather than a $1$. Thus, outputs of the low rank product $\bm{A}^{K,V}_t  \bm{B}^{K,V}_t$ which are less important to learning the new task are zeroed out, leading to precise and minimal changes to the pre-trained weights. Furthermore, since the mask is truly binary (rather than sigmoid), the mask will not learn complex high-rank features (which could potentially interfere with the robust, low-rank fine-tuning properties of $\bm{A}^{K,V}_{t}$ and $\bm{B}^{K,V}_{t}$) and instead provides the model a clear delineation of which parameters are deemed important for the task at hand, as demonstrated in our ablation experiments. The formal loss is given as:
\begin{equation}
\mathcal{L}_{sparse} = || \theta_{\bm{\mathcal{M}}^{K,V}_{t}}(\bm{1}) ||_1
\label{eq:sparse-reg}
\end{equation}

\textbf{MLP for Custom Token Feature Embedding}: 
To further improve the model's customization capability, we replace the custom token feature embeddings $V^*_{t}$ from the previous work with learnable MLP modules $\theta_{V^*_{t}}$, parameterized in the same manner as $\theta_{\bm{\mathcal{M}}^{K,V}_{t}}$. This approach allows the model to adapt the token embeddings based on the specific characteristics of each task, providing a more efficient and flexible way of incorporating task-specific information. While not a key contribution of our work, we found that this custom token parameterization increases the amount of knowledge that can be ``learned'' by the custom tokens, requiring fewer changes to the model and thus less catastrophic interference and forgetting.

\textbf{Putting it all together}: 
Our final optimization is described as:
\begin{equation}
\begin{aligned}
    \underset{\left( \bm{W}^{K,V}_{t} \in \theta, \;\; \theta_{V^*_{t}} \right) }{\operatorname{min}}\mathcal \quad &{L}_{SD}(x,\bm{\theta})+ \lambda_{s} \mathcal{L}_{sparse}(\theta_{\bm{\mathcal{M}}^{K,V}_{t}}(\bm{1})) \\
    &+ \lambda_{f} \mathcal{L}_{forget}(\bm{W}^{K,V}_{t-1},\bm{W}^{K,V}_{t}) 
\end{aligned}
\end{equation}
where $\lambda_s$ is a hyperparameter chosen with a simple exponential sweep. We re-emphasize that, by folding all learned parameters back into the original pre-trained weights, we ensure no additional storage or computational costs at inference, making our approach both storage and compute efficient.

%% file: figures/method.tex
\begin{figure*}[t]
    \centering
    \includegraphics[width=\textwidth]{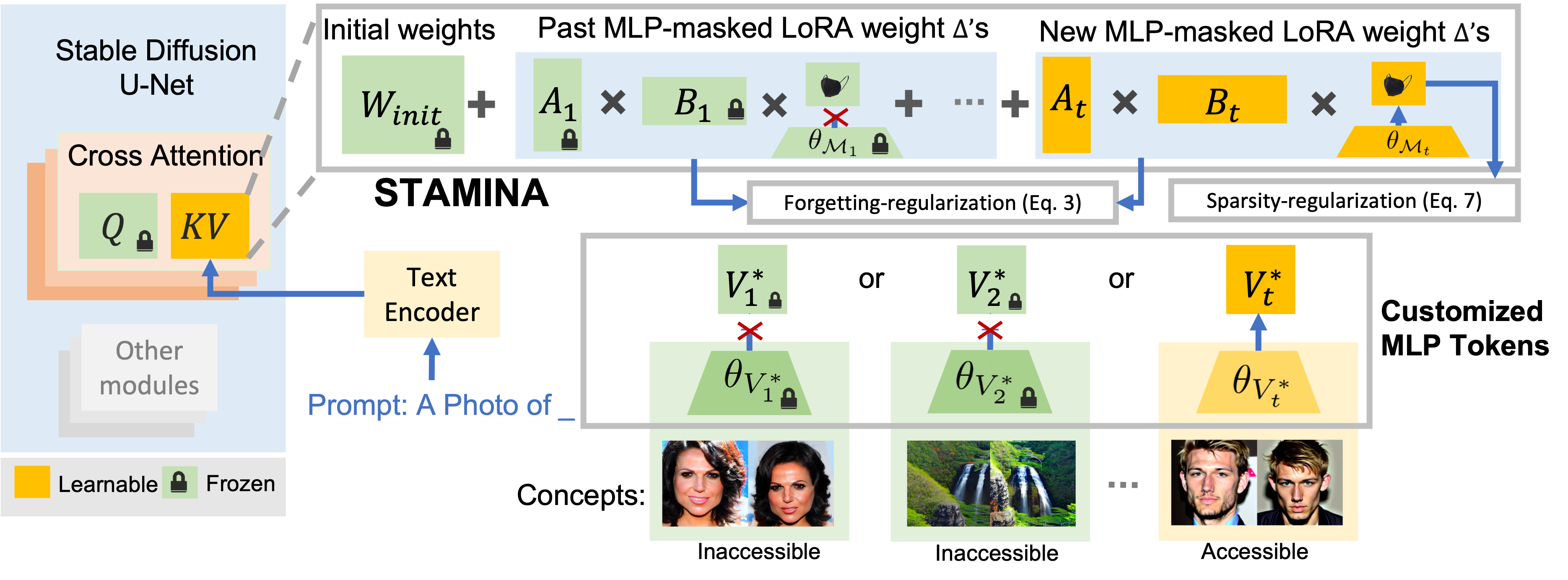}
    \caption{
    An overview of our approach. We learn custom tokens via MLPs operating on a fixed input. A prompt which includes the custom token is passed to the Stable Diffusion model. Our STAMINA approach modifies the key-value (K-V) projection in U-Net cross-attention modules without forgetting by using sparse, low-ranked, adaptations masked with MLP hard-attention. Importantly, trainable parameters, including the MLPs, can be reintegrated back into the original model backbone after training, incurring no cost to storage or inference.
    }
    \vspace{-.5cm}
    \label{fig:method}
\end{figure*}

%% file: sections/5_experiments.tex
\section{Continual Text-To-Image Experiments}
\label{sec:exp_vl}

\textbf{Implementation Details:}
For the most part, we use the same implementation details as Custom Diffusion~\cite{kumari2022multi} and C-LoRA~\cite{smith2023continual}. We use 500 training steps (twice as many as reported in Kumari \emph{et al.}~\cite{kumari2022multi} due to our data being fine-grain concepts rather than simple objects) except for C-LoRA, which requires longer training steps (we use 2000 as reported in Smith \emph{et al.}~\cite{smith2023continual}). We use the prompt ``a photo of a $V*$ X'', where $V*$ is a learnable custom token, and X is the object category (e.g., \emph{person}), which is removed~\cite{smith2023continual} for C-LoRA and STAMINA. For LoRA, we searched for the rank using a simple exponential sweep and found that a rank of 16 sufficiently learns all concepts. Additional training details, including our chosen loss weighting hyperparameter values, are located in \SMa.

\textbf{Metrics:}
We report the same metrics which were originally proposed for \setting~\cite{smith2023continual}: (i) $N_{param}$, the number of parameters \emph{trained} (i.e., unlocked during training a task) in \% of the U-Net backbone model, (ii) $A_{mmd}$, the average MMD score ($\times 10^3$) after training on all concept tasks, and (iii) $F_{mmd}$, average forgetting. Consider $N$ customization ``tasks'' $t \in \{ 1,2,\dots,N-1,N\}$, where $N$ is the total number of concepts that will be shown to our model. We denote $X_{i,j}$ as task $j$ images generated by the model after training task $i$. Furthermore, we denote $X_{D,j}$ as original dataset images for task $j$. Using these terms, we calculate the $A_{mmd}$ metric (where lower is better) as:
\begin{equation}
    A_{mmd} = \frac{1}{N} \sum_{j=1}^{N} MMD \left( \mathcal{F}_{clip}(X_{D,j}),\mathcal{F}_{clip}(X_{N,j}) \right)
\end{equation}
where $\mathcal{F}_{clip}$ denotes a function embedding into a rich semantic space using CLIP~\cite{radford2021learning}, and $MMD$ denotes the Maximum Mean Discrepancy~\cite{gretton2012kernel} with a polynomial kernel. To calculate the forgetting metric $F_{mmd}$, we calculate the average distance the images have changed over training, or:
\begin{equation}
    F_{mmd} = \frac{1}{N-1} \sum_{j=1}^{N-1} MMD \left( \mathcal{F}_{clip}(X_{j,j}),\mathcal{F}_{clip}(X_{N,j}) \right)
\end{equation}

\textbf{Baselines:}
In addition to the prior SOTA C-LoRA~\cite{smith2023continual}, we compare to recent customization methods Textual Inversion~\cite{gal2022image} and Custom Diffusion~\cite{kumari2022multi}. For a better comparison, we compare to a version of Textual Inversion~\cite{gal2022image} which leverages our same MLP custom tokens $\theta_{V^*_{t}}$ rather than $V^*_{t}$, TI++. For Custom Diffusion, we compare to both sequential training (denoted as \emph{sequential}) as well as the constrained merging optimization variant (denoted as \emph{merged}) which stores separate KV parameters for each concept individually and then merges the weights together into a single model using a closed-form optimization (see Kumari \emph{et al.}~\cite{kumari2022multi} for more details). We also compare to the continual learning method EWC~\cite{kirkpatrick2017overcoming} combined with Custom Diffusion. We note here that Generative Replay~\cite{Shin:2017} and DreamBooth~\cite{ruiz2022dreambooth} were found to be non-competitive in this setting~\cite{smith2023continual}, and thus we exclude the results from our experiments.

\subsection{Continual Customization Results}
\input{tables/20-tasks}

\input{figures/results_50}

We first benchmark on two 20-length task sequences: 512x512 resolution celebrity faces dataset, CelebFaces Attributes (Celeb-A) HQ~\cite{karras2017progressive,liu2015deep} (Table~\ref{tab:faces}), and waterfall landmarks of various resolutions from the Google Landmarks dataset v2~\cite{weyand2020google} (Table~\ref{tab:water}). We refer the reader to \SMb ~for additional details on benchmark dataset sampling. We first observe that the Custom Diffusion (CD) methods suffer in this setting, as previously reported by Smith \emph{et al.}~\cite{smith2023continual}. The only two competitive methods in this setting are the prior SOTA C-LoRA~\cite{smith2023continual} and our MLP variant of Textual Inversion (TI)~\cite{gal2022image}, TI++. While TI++ has no forgetting, tokens alone, even when learned using a MLP, do not have the capacity to capture fine-grain details in datasets such as faces and landmarks. We also see that, quantitatively, C-LoRA performs worse than TI in these benchmarks, as explained in Section~\ref{sec:plastic-problem}. On the other hand, we see that our STAMINA approach establishes a clear SOTA performance on both benchmarks, while requiring only 500 training steps (compared to 2000 training steps for C-LoRA).
\input{tables/50-task_ablation}

\textbf{What about longer task sequences?} In Table~\ref{tab:50}, we scale to a 50 length sequence containing \textit{both} celebrity faces~\cite{karras2017progressive,liu2015deep} and landmarks~\cite{weyand2020google}. We note that merge variant of CD is excluded, as it runs into memory errors in the merging process for such a long task sequence. Here, we notice two significant observations: (1) C-LoRA is now performing much worse than TI, while STAMINA is still retaining its SOTA performance. In Figure~\ref{fig:50}, we show qualitative results of images generated for all 50 tasks \emph{after the full task sequence has been seen}. For example, the images in row 1 corresponding to ``task 1'' are being generated from the model after training on ``task 50''. Here, we see quite clearly that our method has gains over the existing techniques. While TI continues to learn decent images of high quality for each task, it misses important identifying details for most tasks. On the other hand, methods such as CD and C-LoRA suffer from catastrophic forgetting and saturated plasticity (which seems to lead to interference and forgetting in early tasks), respectively. We see that our STAMINA method has the best overall results for the full task sequence.

\textbf{Ablations:} We include an ablation study using the 50-concept benchmark in Table~\ref{tab:ablations}. First, we ablate our mask, and see an increase in both forgetting $F_{mmd}$ and MMD score $A_{mmd}$. We observe similar trends when we ablate the sparsity loss \eqref{eq:sparse-reg} and mask MLP parameterization, though with even higher forgetting in these two ablations. This implies it would be better to have no mask then have a ablated versions of our mask. We see a large gap in performance for ``Ablate Gumbel-Softmax'', which is where we use sigmoid activations instead of binary masks. Finally, we see the worst performance when we ablate the MLP Tokens and use $V^*$ tokens instead. We found that increasing the steps can mitigate this to some degree, yet still under-performs our full method. \SMd

\input{figures/multi}

\textbf{Multi-Concept Generations:} In Figure~\ref{fig:multi}, we provide some results demonstrating the ability to generate photos of multiple concepts in the same picture. We found that using the prompt styles ``a photo of V* person posing next to V* waterfall'' worked best. We also provide negative results in \SMc ~and emphasize that there is much room for improvement in future work.

\input{tables/impact}
\textbf{Performance vs Training Time:} Our method shows a $9.6\%$ increase in training time over TI++, due to complex gradient propagation into the token embedding space. Despite this, our approach significantly improves $A_{mmd}$ by $9.1\%$, justifying the trade-off, especially as there's no extra cost during inference. When compared to an \textbf{Upper Bound (UB)} set by CD\cite{kumari2022multi}, \textbf{our method more than halves the performance differential with TI++}, reducing it from $16.67\%$ to $6.01\%$. This highlights our method's efficacy in continual learning, effectively balancing compute efficiency with performance enhancements.

\subsection{\method~for Image Classification}
\label{sec:exp:imnet-r}

\looseness=-1
We also demonstrate that \method~ achieves SOTA performance for a long-task sequence in the well-established setting of \emph{rehearsal-free continual learning for image classification}. We benchmark our approach using the 20 task ImageNet-R~\cite{hendrycks2021many,wang2022dualprompt} benchmark, which is composed of 200 total object classes with a wide collection of image styles, including cartoon, graffiti, and hard examples from the original ImageNet dataset~\cite{russakovsky2015imagenet}. We use the exact same experiment setting as the CODA-Prompt~\cite{smith2022coda} paper with the ViT-B/16 backbone~\cite{dosovitskiy2020image} pre-trained on ImageNet-1K~\cite{russakovsky2015imagenet} (additional details are available in \SMe). We compare to Learning to Prompt (L2P)~\cite{wang2022learning},  DualPrompt~\cite{wang2022dualprompt}, CODA-Prompt~\cite{smith2022coda} (CODA-P), and C-LoRA~\cite{smith2023continual}. For \method~and C-LoRA, we modify only the QKV projection matrices of self-attention blocks throughout the ViT model.
\input{figures/imnet}

In Figure~\ref{fig:imnet-r}, we plot average accuracy $A_{n}$, or the accuracy with respect to all past classes averaged over all seen $n$ tasks, for all 20 tasks. We choose the 20 task benchmark over the 5 and 10 task benchmarks because C-LoRA performs \emph{under SOTA} in this longer task sequence, as reported in Smith \emph{et al.}~\cite{smith2023continual} (and is consistent with our \setting~findings). Our results demonstrate that \method~performs much better than C-LoRA, and \emph{establishes a new SOTA performance on the benchmark} by outperforming all three prompting-based methods.

%% file: tables/20-tasks.tex
\begin{table}[t]
\caption{\textbf{20-task \setting~Results:} $A_{mmd}$ ($\downarrow$) gives the average MMD score ($\times 10^3$) after training on all concept tasks, and $F_{mmd}$ ($\downarrow$) gives the average forgetting. $N_{param}$ ($\downarrow$) gives the number of parameters being trained as a  \% of the unmodified U-Net backbone size.
}
\vspace{-0.2cm}
\begin{subtable}{0.49\textwidth}
\caption{Celeb-A HQ~\cite{karras2017progressive,liu2015deep}}
\label{tab:faces}
\centering

\begin{tabular}{c c ||c c} 
\hline 
Method & \thead{$N_{param}$ \\Train ($\%$)} & $A_{mmd}$ ($\downarrow$) & $F_{mmd}$ ($\downarrow$) \\
\hline
TI++~\cite{gal2022image}  & $0.00$ & $2.37$ & $\bm{0.00}$ \\
CD~\cite{kumari2022multi}  & $2.23$  & $7.58$ & $6.56$ \\
CD~\cite{kumari2022multi} (Merge) & $2.23$  & $13.84$ & $8.61$ \\
CD+EWC~\cite{kirkpatrick2017overcoming}  & $2.23$  & $7.39$ & $5.81$ \\
C-LoRA~\cite{smith2023continual} & $0.09$ & $2.25$ & $0.33$ \\
\hline
Ours & $0.19$ & $\bm{2.18}$ & $0.03$ \\
\hline
\end{tabular}

\end{subtable}
\hfill
\begin{subtable}{0.49\textwidth}
\vspace{.5cm}
\caption{Google Landmarks dataset v2~\cite{weyand2020google}}
\label{tab:water}
\centering

\begin{tabular}{c c ||c c} 
\hline 
Method & \thead{$N_{param}$ \\Train ($\%$)} & $A_{mmd}$ ($\downarrow$) & $F_{mmd}$ ($\downarrow$) \\
\hline
TI++~\cite{gal2022image}  & $0.00$ & $2.91$ & $\bm{0.00}$ \\
CD~\cite{kumari2022multi}  & $2.23$ & $5.20$ & $5.10$ \\
CD~\cite{kumari2022multi} (Merge) & $2.23$ & $14.83$ & $8.43$ \\
CD+EWC~\cite{kirkpatrick2017overcoming}  & $2.23$ & $5.10$ & $3.56$ \\
C-LoRA~\cite{smith2023continual} & $0.09$ & $3.09$ & $0.38$ \\
\hline
Ours & $0.19$ & $\bm{2.42}$ & $0.01$ \\
\hline
\end{tabular}

\end{subtable}
\vspace{-.1cm}
\end{table}

%% file: figures/results_50.tex
\begin{figure*}[!ht]
    \centering
    \begin{subfigure}[t]{0.19\textwidth}
        \centering
        \includegraphics[width=1\textwidth]{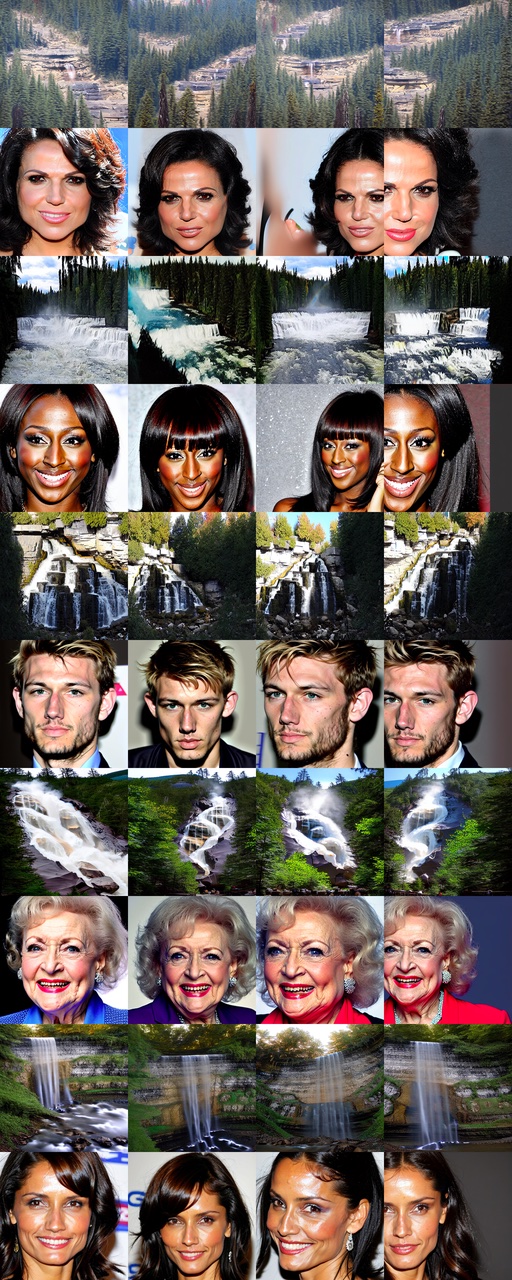}
         \caption{\red{Target}}
    \end{subfigure}
    \hfill
    \begin{subfigure}[t]{0.19\textwidth}
        \centering
        \includegraphics[width=1\textwidth]{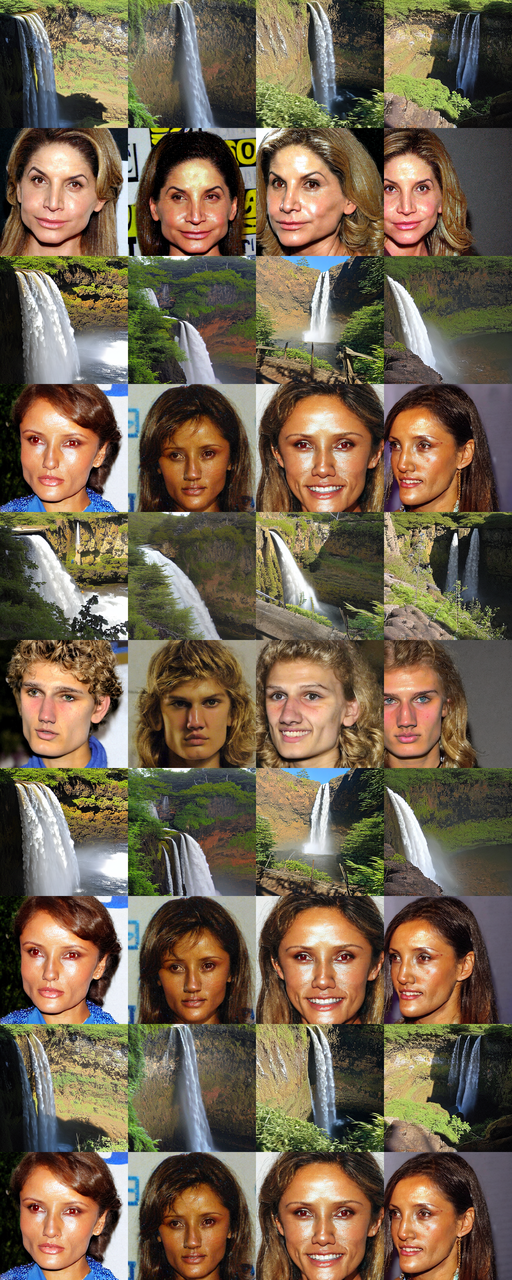}
         \caption{\blue{CD}~\cite{kumari2022multi}}
    \end{subfigure}
    \hfill
    \begin{subfigure}[t]{0.19\textwidth}
        \centering
        \includegraphics[width=1\textwidth]{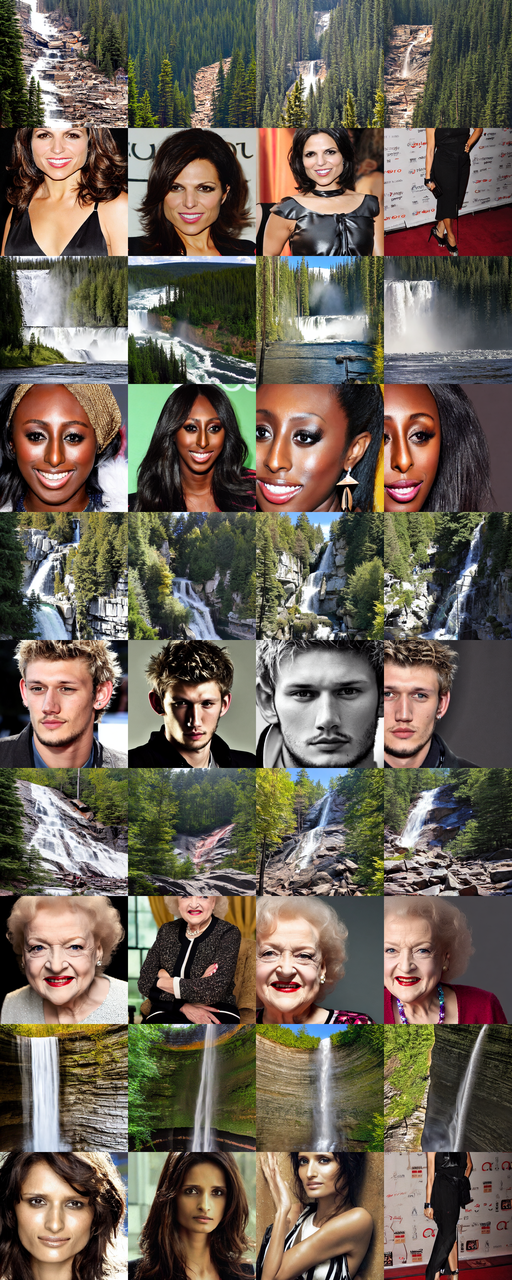}
         \caption{\blue{TI++}~\cite{gal2022image}}
    \end{subfigure}
    \hfill
    \begin{subfigure}[t]{0.19\textwidth}
        \centering
        \includegraphics[width=1\textwidth]{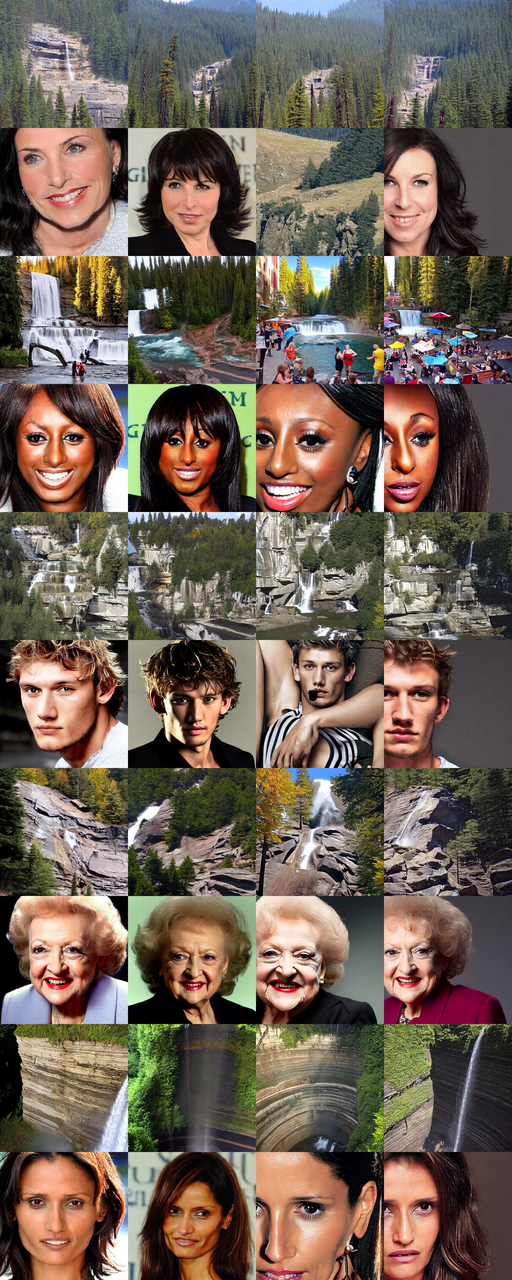}
         \caption{\blue{C-LoRA}~\cite{smith2023continual}}
    \end{subfigure}
    \hfill
    \begin{subfigure}[t]{0.19\textwidth}
        \centering
        \includegraphics[width=1\textwidth]{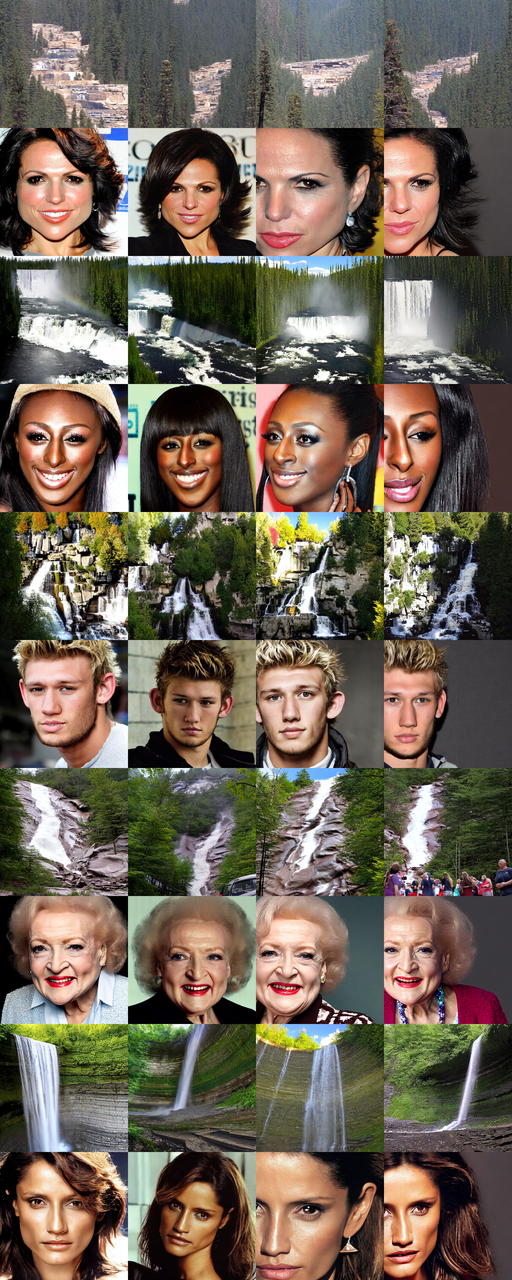}
         \caption{\blue{STAMINA}}
    \end{subfigure}
    \caption{
    Qualitative results of \setting~using celebrity faces from Celeb-A HQ~\cite{karras2017progressive,liu2015deep} and waterfalls from Google Landmarks~\cite{weyand2020google}. Results are shown for 10 samples from all 50 concepts ($\downarrow$) and are \textbf{generated from the model \emph{after} training on all 50 concepts}. We sample for a variety of early (prone to forgetting) and late (prone to low plasticity) tasks. \SMf
    }
    \label{fig:50}
    \vspace{-.2cm}
\end{figure*}

%% file: tables/50-task_ablation.tex
\begin{table}[t]
\caption{\textbf{50-task \setting~Results: } 
$A_{mmd}$ ($\downarrow$) gives the average MMD score ($\times 10^3$) after training on all concept tasks, and $F_{mmd}$ ($\downarrow$) gives the average forgetting. $N_{param}$ ($\downarrow$) gives the number of parameters being trained as a  \% of the unmodified U-Net backbone size.
}
\vspace{-0.2cm}
\begin{subtable}{0.49\textwidth}
\caption{Full results}
\label{tab:50}
\centering
\begin{tabular}{c c ||c c} 
\hline 
Method & \thead{$N_{param}$ \\Train ($\%$)} & $A_{mmd}$ ($\downarrow$) & $F_{mmd}$ ($\downarrow$) \\
\hline
TI++~\cite{gal2022image}  & $0.00$ & $2.52$ & $\bm{0.00}$ \\
CD~\cite{kumari2022multi} &  $2.23$  & $5.99$ & $5.67$ \\
CD+EWC~\cite{kirkpatrick2017overcoming} & $2.23$  & $5.15$ & $3.95$ \\
C-LoRA~\cite{smith2023continual} & $0.09$  & $3.09$ & $1.41$ \\
\hline
Ours & $0.19$ & $\bm{2.29}$ & $0.01$ \\
\hline
\end{tabular}

\end{subtable}
\hfill
\begin{subtable}{0.49\textwidth}
\vspace{.5cm}
\caption{Ablation results}
\label{tab:ablations}
\centering

\begin{tabular}{c ||c c} 
\hline 
Ablation & $A_{mmd}$ ($\downarrow$) & $F_{mmd}$ ($\downarrow$) \\
\hline
Full Method & $\bm{2.29}$ & $\bm{0.01}$ \\
\hline
Ablate Mask & $2.91$ & $0.20$ \\
Ablate MLP Tokens & $3.89$ & $0.29$ \\
Ablate Mask MLP & $2.82$ & $0.58$ \\
Ablate Gumbel-Softmax & $3.39$ & $0.82$ \\
Ablate Sparsity & $2.90$ & $0.56$ \\
\hline

\end{tabular}
\vspace{-.3cm}
\end{subtable}
\end{table}

%% file: figures/multi.tex
\begin{figure}[t]
    \centering
    \includegraphics[width=.45\textwidth,scale=0.9]{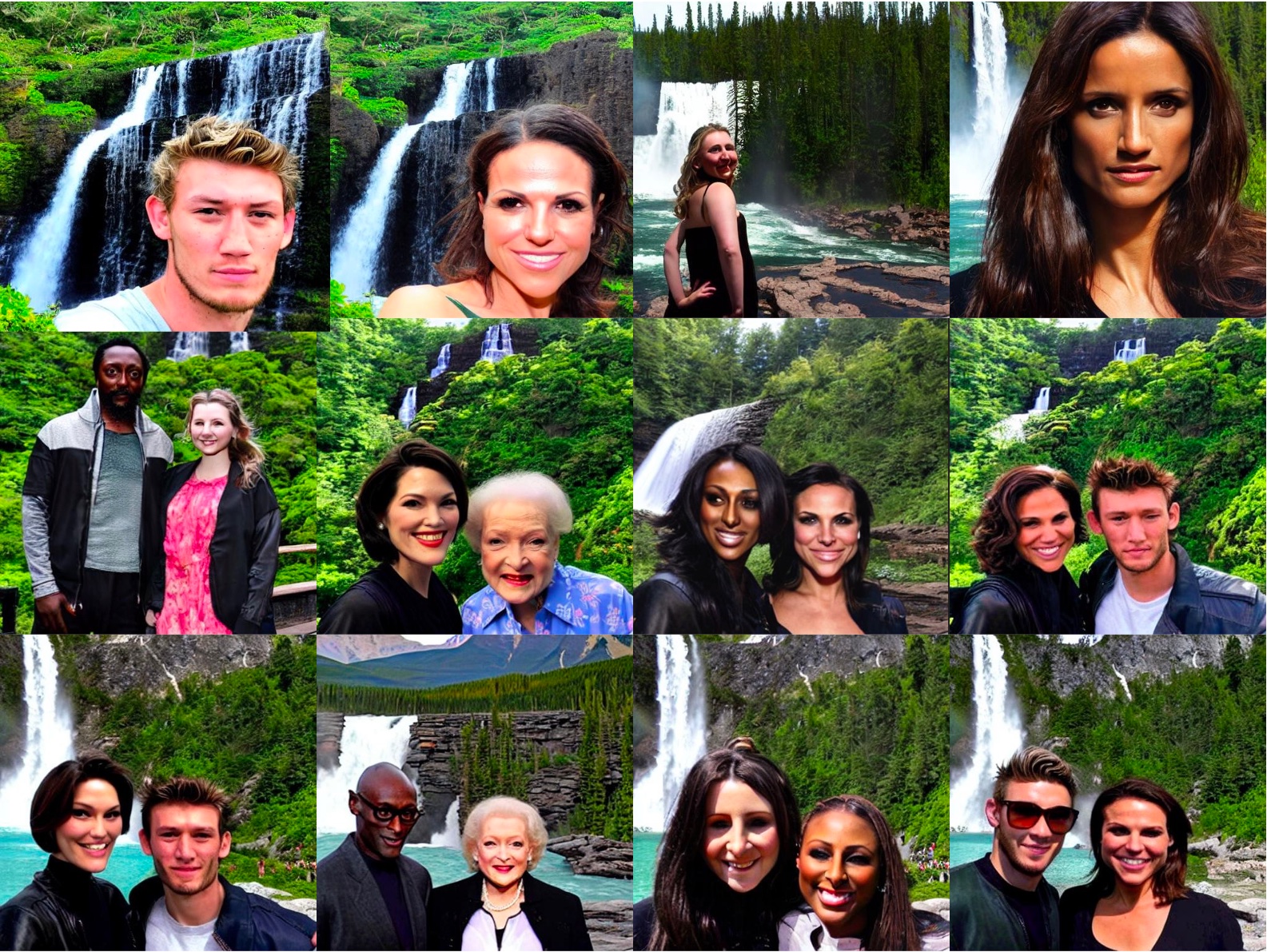}
    \caption{Our multi-concept generations after training on 50 tasks.}
    \label{fig:multi}
    \vspace{-.1cm}
\end{figure}

%% file: tables/impact.tex
\begin{table}[t]
\caption{\textbf{Impact of our Results (50-Task Sequence)}. UB stands for Upper Bound performance. 
}
\centering
\label{tab:impact}
\begin{tabular}{c || c c c} 
\hline
Method       & TI++~\cite{gal2022image}  & Ours & UB \\ \hline
$A_{mmd}$ ($\downarrow$)       & $2.52$  & $\bm{2.29}$       & $2.16$             \\ \hline
$\Delta$ from UB     & $0.36$  & $\bm{0.13}$       & $0$                \\ \hline
\% $\uparrow$ from UB          & $16.67\%$ & $\bm{6.02\%}$    & $0\%$              \\ \hline
\end{tabular}
\vspace{-.2cm}
\end{table}

%% file: figures/imnet.tex
\begin{figure}[t]
    \centering
    \includegraphics[width=.42\textwidth]{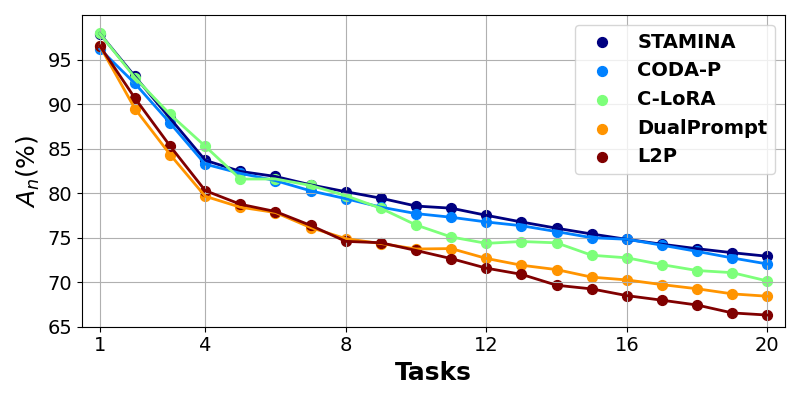}
     \caption{Image classification results on ImageNet-R~\cite{hendrycks2021many,wang2022dualprompt}. $A_{n}$ gives the accuracy averaged over all seen tasks.}
     \label{fig:imnet-r}
    \vspace{-.35cm}
\end{figure}

%% file: sections/6_conclusions.tex
\section{Conclusion \& Limitations}
\label{sec:conclusion}

\looseness=-1
In conclusion, our work addresses the challenge of scaling text-to-image diffusion models for continual customization across long concept sequences. We showed that the the SOTA C-LoRA's performance saturates with increasing tasks, and proposed STAMINA as a novel and efficient solution. STAMINA is composed of low-rank adapters with sparse, hard-attention masking and learnable MLP tokens, and can reintegrate all trainable parameters back into the original model backbone post-training. Our comprehensive evaluations on 3 different \setting~benchmarks not only highlight STAMINA's superior performance over the current SOTA, but also its efficiency, requiring significantly fewer training steps. Furthermore, we show that \method~is also SOTA for image classification, demonstrating its flexibility to achieve high performance in multiple continual learning settings.

\looseness=-1
Despite the success of our approach in generating long concept sequences, we acknowledge key limitations and cautions that must be addressed. \textbf{First, we strongly advocate for the responsible usage of our approach}, particularly with regards to generating faces of individuals. Consent, in our view, is paramount. Furthermore, given ethical concerns over using artists' and designers' images, we avoid artistic creativity in our work, and urge others to know and respect the sources of the data in which they customize their model with. In spite of these ethical considerations, we remain optimistic about the potential of our work to contribute positively to society in use cases such as mobile app entertainment.

%% file: sections/7_appendix_arxiv.tex
\section{Analysis on Interference}
\input{figures/sm_analysis}
\label{app:interference_analysis}

In Figure~\ref{fig:sm:perc-train-new}, we show that \method~ has \emph{low interference} in changes to the pre-trained weights over tasks. Specifically, we plot the percentage of non-zero $\bm{W}^{K,V}_{t}-\bm{W}^{K,V}_{t-1}$ weight adaptations which are modifying the pre-trained weights $\bm{W}^{K,V}_{init}$ in their corresponding locations (i.e., indices in the weight matrix) for the first time. The reader should recall that our weight adaptations are \emph{sparse} due to a hard masking mechanism (Eq.~\ref{eq:stamina}) and sparsity regularization loss (Eq.~\ref{eq:sparse-reg}). Thus, in combination with the forgetting loss (Eq.~\ref{eq:lora-reg}), our method should naturally \emph{avoid altering the pre-trained weights in the same index locations as previous tasks}. We show this exactly - over 50 tasks, the percentage remains high, indicating little to no interference during each task. We note that in some tasks the percentage drops below 100\%, demonstrating that some interference still exists in our method.

On the contrary, this same plot for C-LoRA~\cite{smith2023continual} and Custom Diffusion~\cite{kumari2022multi} would, by the designs of these methods, show close to or exactly 0\% from tasks 2 and beyond, indicating \emph{high interference} at each task. This high interference is likely a strong contributor to the increased catastrophic forgetting of past task concepts in these methods.

\section{Additional Metrics}
\input{tables/sm-new-metrics}
\label{app:metrics}

In the main paper tables, we provided the following metrics: $A_{mmd}$ ($\downarrow$), which gives the average MMD score ($\times 10^3$) after training on all concept tasks, $F_{mmd}$ ($\downarrow$), which gives the average forgetting, and $N_{param}$ ($\downarrow$), which gives the \% number of parameters being trained. To provide additininal context to our experiments, we provide: KID ($\downarrow$), which gives the Kernel Inception Distance ($\times 10^3$) between generated and dataset images, and plasticity $P_{mmd}$ ($\downarrow$), which gives the average plasticity (ability to learn new tasks) as the average MMD score ($\times 10^3$) for all concepts measured directly after after training. The new metrics can be found in Tables~\ref{tab:sm:50},\ref{tab:sm:water},\ref{tab:sm:faces}.
\vspace{-0.6em}
\begin{equation}
    P_{mmd} = \frac{1}{N} \sum_{j=1}^{N} MMD \left( \mathcal{F}_{clip}(X_{D,j}),\mathcal{F}_{clip}(X_{j,j}) \right)
\end{equation}

\section{Plasticity Analysis}
\input{figures/plastic-analyze}
\label{app:plastic_analysis}

In Figure~\ref{fig:sm:plastic}, we directly compares plasticity vs. number of trained tasks for C-LoRA, TI++, and STAMINA in the Table~\ref{tab:50} 50 task benchmark. This figure shows (i) a stronger decrease in plasticity for C-LoRA (compared to STAMINA) and (ii) C-LoRA converging to a much worse plasticity value.

\section{Additional Implementation Details}
\input{figures/sm_multi}
\label{app:implementation_details}

\looseness=-1
We use 2 A100 GPUs to generate all results. All hyperparameters were searched with an exponential search (for example, learning rates were chosen in the range $5e-2,5e-3,5e-4,5e-5,5e-6,5e-7,5e-8$). We found a learning rate of $5e-6$ worked best for the Custom Diffusion~\cite{kumari2022multi} methods, and a learning rate of $5e-4$ worked best for the LoRA-based methods and Textual Inversion~\cite{gal2022image}. Following Smith \emph{et al.}~\cite{smith2023continual}, we use a loss weight of $1e6$ and $1e8$ for EWC~\cite{kirkpatrick2017overcoming} and C-LoRA, respectively. For our method, we found a loss weight of $1e-3$ and $1e3$ worked best for the sparsity penalty (Eq.~\ref{eq:sparse-reg}) and forgetting loss (Eq.~\ref{eq:lora-reg}), respectively. We found a rank of 16 was sufficient for LoRA for the text-to-image experiments and 64 for the image classification experiments. These were chosen from a range of $8,16,32,64,128$. We use 500 training steps (twice as many as reported in Kumari \emph{et al.}~\cite{kumari2022multi} due to our data being fine-grain concepts rather than simple objects) except for C-LoRA, which requires longer training steps (we use 2000 as reported in Smith \emph{et al.}~\cite{smith2023continual}). We regularize training with generated auxiliary data (as done in Smith \emph{et al.}~\cite{smith2023continual}) for \emph{all} methods.

The simple MLPs used in our paper are composed of two linear layers and a ReLU~\cite{agarap2018deep} layer in between. For the mask MLPs, $\theta_{\bm{\mathcal{M}}^{K,V}_{t}}$, the dimension of linear layers 1 and 2 are $r \times r$ and $r \times D_1 \cdot D_2 \cdot 2$, where $r$ is the same low rank as the LoRA parameters $\bm{A}^{K,V}_{t}$ and $\bm{B}^{K,V}_{t}$, and $D1,D2$ are the dimensions of the weight $\bm{W}^{K,V}$. For the custom token MLPs $\theta_{V^*_{t}}$, the dimension of linear layers 1 and 2 are both $D_{token} \times D_{token} $, where $D_{token}$ is the dimension of the token embedding.
 
\section{Benchmark Dataset Details}
\label{app:benchmark_details}
Given the datasets Celeb-A HQ~\cite{karras2017progressive,liu2015deep} and Google Landmarks v2~\cite{weyand2020google}, we sample concepts at random which have at least 10 individual training images each. Specifically, we iterate randomly over the fine-grained identities of each dataset (person for Celeb-A HQ and waterfall location for Google Landmarks V2) and check whether the identity has sufficient unique examples in the dataset; we do this until we reached the number of desired concepts for each dataset. Each concept customization is considered a ``task'', and the tasks are shown to the model sequentially.

\section{Additional Details for Image Classification Setting}
\label{app:details_classification}
In Section 5.2, we benchmark our approach using ImageNet-R~\cite{hendrycks2021many,wang2022dualprompt} which is composed of 200 object classes with a wide collection of image styles, including cartoon, graffiti, and hard examples from the original ImageNet dataset~\cite{russakovsky2015imagenet}. This benchmark is chosen because the distribution of training data has significant distance to the pre-training data (ImageNet), thus providing a problem setting which is both fair and challenging.

We use the same experimental settings as those used in the recent CODA-Prompt~\cite{smith2022coda} paper. We implement our method and all baselines in PyTorch\cite{paszke2019pytorch} using the ViT-B/16 backbone~\cite{dosovitskiy2020image} pre-trained on ImageNet-1K~\cite{russakovsky2015imagenet}. All methods are trained with a batch size of 128 for 50 epochs; the prompting-based methods use a learning rate of $5e-3$, whereas the LoRA based methods use a learning rate of $5e-4$. We compare to the following methods (the same rehearsal-free comparisons of CODA-Prompt): CODA-Prompt~\cite{smith2022coda}, Learning to Prompt (L2P)~\cite{wang2022learning}, DualPrompt~\cite{wang2022dualprompt}, and C-LoRA~\cite{smith2023continual}. We use the same classification head as L2P, DualPrompt, and CODA-Prompt. For additional details, we refer the reader to original CODA-Prompt~\cite{smith2022coda} paper. For our method, we add \method~to the QKV projection matrices of self-attention blocks throughout the ViT model, and use the same 64 rank as used in C-LoRA~\cite{smith2023continual}.

\section{Negative Multi-Concept Results}
\label{app:negative}

\looseness=-1
We extend our results demonstrating the ability to generate photos of multiple concepts in the same picture by showing both successful attempts (Figure~\ref{fig:sm:multi-good}) and failing attempts (Figure~\ref{fig:sm:multi-bad}). We use the prompt style ``a photo of V* person posing next to V* waterfall'' for the top row (single person and single landmark) and ``a photo of V* person, standing next to V* person, posing in front of V* waterfall'' for rows 2 and 3 (two people and a single landmark). Unlike most results in our paper, which diffuse for 200 steps (as done in \cite{kumari2022multi}), we allow the multi-concept results to diffuse for 500 steps.

Each generated image in Figure~\ref{fig:sm:multi-bad} used the same prompt as the corresponding image in Figure~\ref{fig:sm:multi-good}. In general, we found a success rate of roughly 50\% for two concept generations and 20\% for the challenging 3 concept generations. The failures in row 1 (single person with single landmark) each have a blurred or occluded concept. In rows 2 and 3 (two people with single landmark), we see failures such as the landmark disappearing (row 2, column 1), imagined people (row 2, column 4), merged people (row 3, column 2), or one concept taking on characteristics of another person, such as skin tone (row 3, column 3) or age (row 2, column 2), \emph{which could be explained by bias and is a limitation that users of this work should pay close attention to.} We hope to address these sources of failures in future work.

\section{Variance Across Runs}
\label{app:variance}
\input{tables/SM_error-bars}
In Table~\ref{tab:sm:error}, we provide the mean and standard deviation for each method across all 3 \setting~benchmarks (Tables~\ref{tab:faces}, \ref{tab:water}, and \ref{tab:50}). We see that our method not only has the best metric performance, but also has the lowest standard deviation for both $A_{mmd}$ and $F_{mmd}$.

\section{Figure Image Sources}
\label{appendix:source}
In our figures, we replace dataset images with generated similar images due to licensing constraints. Specifically, we generate ``target data'' using offline (i.e., no \emph{continual} learning) single-concept Custom Diffusion~\cite{kumari2022multi}, which we refer to as \emph{pseudo figure images}. We note that all training and evaluations were completed using the original datasets, and all result images were obtained through models trained directly on the original datasets. For Figure~\ref{fig:use-case}, the images captioned ``learn'' are \emph{pseudo figure images}, and the multi-concept images are results produced with our method. For Figure~\ref{fig:method}, all concept images are \emph{pseudo figure images}. For Figure~\ref{fig:50}, the images labeled ``target data'' are \emph{pseudo figure images}, and the rest are results from models we trained. Finally, Figures~\ref{fig:multi}, \ref{fig:sm:multi-good}, and \ref{fig:sm:multi-bad} only contain results produced from models we trained.

%% file: figures/sm_analysis.tex
\begin{figure}[h]
    \centering
    \includegraphics[width=.47\textwidth]{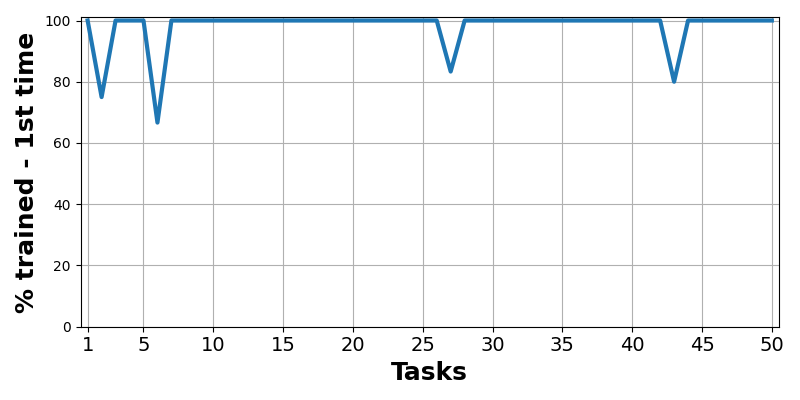}
     \caption{Percentage of non-zero $\bm{W}^{K,V}_{t}-\bm{W}^{K,V}_{t-1}$ adaptations which are modifying the pre-trained weights $\bm{W}^{K,V}_{init}$ at the corresponding position for the first time. Here, a high number equates to low interference (good), and a low number equates to high interference (bad).}
     \label{fig:sm:perc-train-new}
\end{figure}

%% file: tables/sm-new-metrics.tex
\begin{table*}[h]
\caption{50-Task Full Results}
\label{tab:sm:50}
\centering
\begin{tabular}{c c ||c c c c} 
\hline 
Method & \thead{$N_{param}$ \\Train ($\%$)} & $A_{mmd}$ ($\downarrow$) & $F_{mmd}$ ($\downarrow$) & KID ($\downarrow$) & $P_{mmd}$ ($\downarrow$) \\
\hline
TI++ [2]  & $0.00$ & $2.52$ & $\bm{0.00}$ & $38.33$ & $2.56$ \\
CD [3] &  $2.23$  & $5.99$ & $5.67$ & $85.08$ & $3.17$ \\
CD+EWC [21] & $2.23$  & $5.15$ & $3.95$ & $64.47$ & $3.45$ \\
C-LoRA [1] & $0.09$  & $3.09$ & $1.41$ & $45.37$ & $2.79$ \\
\hline
Ours & $0.19$ & $\bm{2.29}$ & $0.01$ & $\bm{25.73}$ & $\textbf{2.32}$ \\
\hline
\end{tabular}
\end{table*}
\begin{table*}[h]
\caption{20-Task Results on Google Landmarks dataset v2 [59]}
\label{tab:sm:water}
\centering
\begin{tabular}{c c ||c c c c} 
\hline 
Method & \thead{$N_{param}$ \\Train ($\%$)} & $A_{mmd}$ ($\downarrow$) & $F_{mmd}$ ($\downarrow$) & KID ($\downarrow$) & $P_{mmd}$ ($\downarrow$) \\
\hline
TI++ [2]  & $0.00$ & $2.91$ & $\bm{0.00}$ & $33.69$ & $3.03$ \\
CD [3]  & $2.23$ & $5.20$ & $5.10$ & $114.55$ & $3.25$ \\
CD [3] (Merge) & $2.23$ & $14.83$ & $8.43$ & $331.21$ & $10.19$ \\
CD+EWC [21]  & $2.23$ & $5.10$ & $3.56$ & $80.58$ & $3.23$ \\
C-LoRA [1] & $0.09$ & $3.09$ & $0.38$ & $53.24$ & $3.15$ \\
\hline
Ours & $0.19$ & $\bm{2.42}$ & $0.01$ & $\bm{31.73}$ & $\textbf{2.44}$ \\
\hline
\end{tabular}
\end{table*}
\begin{table*}[h!]
\caption{20-Task Results on Celeb-A HQ [57,58]}
\label{tab:sm:faces}
\centering
\begin{tabular}{c c ||c c c c} 
\hline 
Method & \thead{$N_{param}$ \\Train ($\%$)} & $A_{mmd}$ ($\downarrow$) & $F_{mmd}$ ($\downarrow$) & KID ($\downarrow$) & $P_{mmd}$ ($\downarrow$) \\
\hline
TI++ [2]  & $0.00$ & $2.37$ & $\bm{0.00}$ & $35.49$ & $2.35$ \\
CD [3]  & $2.23$  & $7.58$ & $6.56$ & $104.54$ & $3.43$ \\
CD [3] (Merge) & $2.23$  & $13.84$ & $8.61$ & $353.40$ & $7.83$ \\
CD+EWC [21]  & $2.23$  & $7.39$ & $5.81$ & $91.61$ & $3.45$ \\
C-LoRA [1] & $0.09$ & $2.25$ & $0.33$ & $37.41$ & $2.15$ \\
\hline
Ours & $0.19$ & $\bm{2.18}$ & $0.03$ & $\bm{28.63}$ & $\bm{2.07}$ \\
\hline
\end{tabular}
\end{table*}

%% file: figures/plastic-analyze.tex
\begin{figure}[h!]
    \centering
    \includegraphics[width=.45\textwidth]{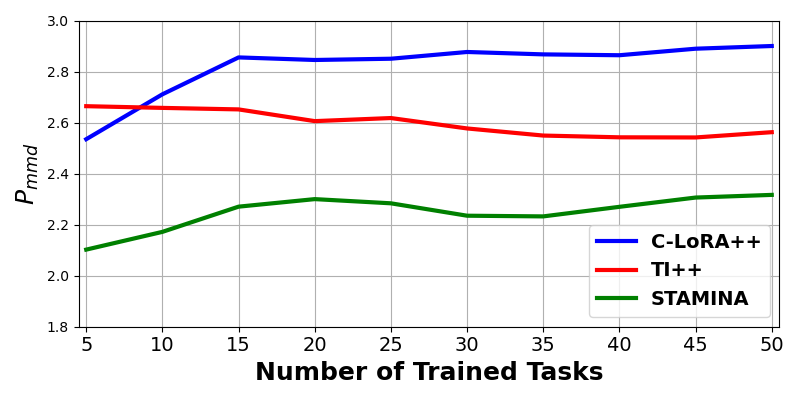}
     \caption{Average plasticity $P_{mmd}$ ($\downarrow$) vs. number of trained tasks.}
     \label{fig:sm:plastic}

\end{figure}

%% file: figures/sm_multi.tex
\begin{figure*}[t]
    \centering
    \begin{subfigure}[t]{0.49\textwidth}
        \centering
        \includegraphics[width=\textwidth]{figures/source/multi.jpg}
         \caption{Successes}
         \label{fig:sm:multi-good}
    \end{subfigure}
    \hfill
    \begin{subfigure}[t]{0.49\textwidth}
        \centering
        \includegraphics[width=1\textwidth]{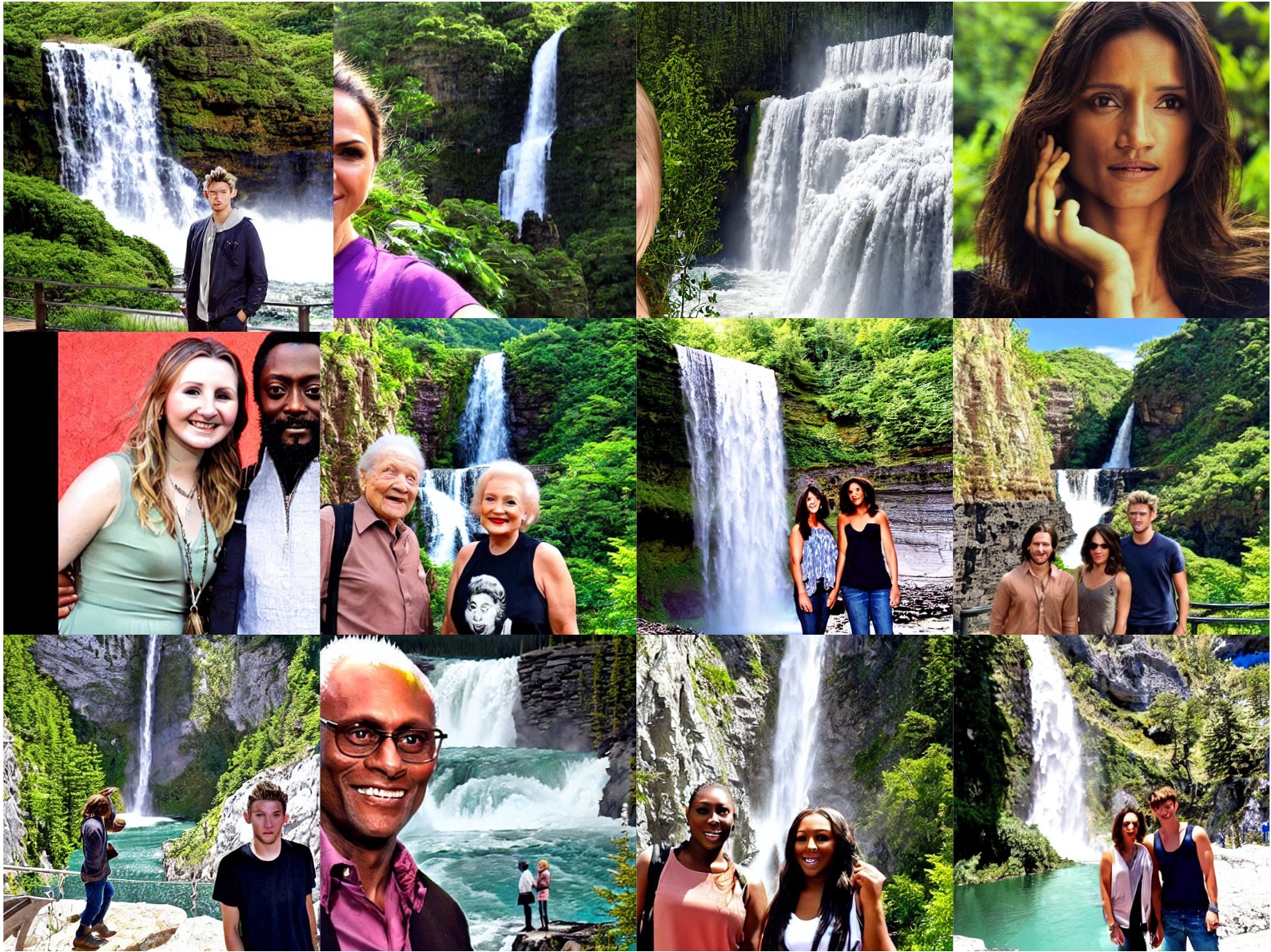}
         \caption{Failures}
         \label{fig:sm:multi-bad}
    \end{subfigure}
    \caption{\method~multi-concept generations after training on 50 tasks.}
\end{figure*}

%% file: tables/SM_error-bars.tex
\begin{table}[h]
\caption{\textbf{Mean and standard deviation across 3 runs: } $A_{mmd}$ ($\downarrow$) gives the average MMD score ($\times 10^3$) after training on all concept tasks, and $F_{mmd}$ ($\downarrow$) gives the average forgetting. $N_{param}$ ($\downarrow$) gives the number of parameters being trained as a  \% of the unmodified U-Net backbone size.
}
\caption{Celeb-A HQ~\cite{karras2017progressive,liu2015deep}}
\label{tab:sm:error}
\centering
\resizebox{0.47\textwidth}{!}{
\begin{tabular}{c c ||c c} 
\hline 
Method & \thead{$N_{param}$ \\Train ($\%$)} & $A_{mmd}$ ($\downarrow$) & $F_{mmd}$ ($\downarrow$) \\
\midrule
TI++~\cite{gal2022image} & 0.00 & 2.60 $\pm$ 0.23 & \textbf{0.00 $\pm$ 0.00} \\
CD~\cite{kumari2022multi} & 2.23 & 6.26 $\pm$ 0.99 & 5.78 $\pm$ 0.60 \\
CD~\cite{kumari2022multi} (Merge) & 2.23 & 14.34 $\pm$ 0.50 & 8.52 $\pm$ 0.09\\
CD+EWC~\cite{kirkpatrick2017overcoming} & 2.23 & 5.88 $\pm$ 1.07 & 4.44 $\pm$ 0.98 \\
C-LoRA~\cite{smith2023continual} & 0.09 & 2.81 $\pm$ 0.40 & 0.71 $\pm$ 0.50 \\
\hline 
Ours & 0.19 & \textbf{2.30 $\pm$ 0.10} & 0.02 $\pm$ 0.01 \\
\hline 
\end{tabular}
}
\end{table}